\documentclass[conference,a4paper]{IEEEtran}
\IEEEoverridecommandlockouts

\usepackage[hidelinks]{hyperref}
\usepackage[cmex10]{amsmath}
\usepackage{amssymb,amsfonts}
\usepackage{dblfloatfix}

\usepackage[ruled,vlined]{algorithm2e}
\usepackage{graphicx}
\graphicspath{{Figures/PDF/}{Figures/PNG/}}

\usepackage{booktabs}
\usepackage{siunitx}
\usepackage[numbers,compress]{natbib}
\usepackage{texnames}
\usepackage{bm,bbm}
\usepackage{orcidlink}

\begin{document}

\title{\uppercase{TempTPI: Informer-Based trajectory prediction for maritime vessels}
}

\author{
	\IEEEauthorblockN{
        Kevin Ferneding\orcidlink{0009-0000-9243-4313},
        Veronika Lietavcova\orcidlink{0009-0005-5037-6553},
        Aleksandra M. Blachowiak\orcidlink{0009-0003-9308-0731},
        Peder Heiselberg\orcidlink{0000-0002-8847-634X}
        }
\IEEEauthorblockA{
\textit{Technical University of Denmark}\\
Anker Engelunds Vej 101
2800 Kongens Lyngby, Denmark\\
ph@space.dtu.dk
}
}

\maketitle
\begin{tikzpicture}[remember picture, overlay]
  \node[anchor=south, yshift=15pt] at (current page.south) {
    \fbox{\parbox{\dimexpr0.88\textwidth-2\fboxsep-2\fboxrule\relax}{
      \footnotesize\centering © 2026 IEEE. Personal use of this material is permitted. 
      Permission from IEEE must be obtained for all other uses, in any current or future media, 
      including reprinting/republishing this material for advertising or promotional purposes, 
      creating new collective works, for resale or redistribution to servers or lists, 
      or reuse of any copyrighted component of this work in other works.
    }}
  };
\end{tikzpicture}

\begin{abstract}
Accurate long-term trajectory prediction for maritime vessels is essential for safety and logistical efficiency. While deep learning models, particularly Transformers, have shown promise in processing Automatic Identification System (AIS) data, they often struggle with the quadratic computational complexity of self-attention and the loss of accuracy over extended forecasting horizons. This study proposes TempTPI, a novel prediction framework that integrates an Informer-based encoder with a multi-channel temporal encoding mechanism. The Informer architecture leverages a ProbSparse self-attention mechanism to reduce computational overhead and focus on the most significant dependencies, while the temporal encoder utilizes Fourier-like frequency expansions to capture cyclic patterns (hourly, daily, and seasonal) in vessel behavior. We evaluate our model against the state-of-the-art TPTrans architecture using AIS data from Danish waters. Experimental results demonstrate that TempTPI consistently outperforms existing methods across prediction windows of 1 to 5 hours. Notably, at a 5-hour horizon, the proposed model achieves a 55\% improvement in Mean Squared Error (MSE), offering a robust solution for long-range maritime situational awareness.
\end{abstract}


\section{Introduction}
\label{sec:intro}


Maritime cargo vessels make up a large portion of global trade \cite{seabornetrade} and require precise orchestration to run flawlessly. 
Large numbers of simultaneous maritime vessels in limited space logically lead to an increase in complexity for securing safe and efficient transportation \cite{maritimescenario}.
Thus, the potential commercial benefits of a high-accuracy trajectory prediction for commercial vessels in order to enable a reliable vessel coordination are very high \cite{benefits,shiptracking}. 

A widely used communication system in the maritime domain is the \textit{Automatic Identification System} (AIS), which aims to transmit real-time information about the current status of the ship \cite{reviewofais} for collision avoidance. Through the wide distribution of this system, a significant amount of data has been collected over time, which provides a valuable foundation for deep learning research. The Danish Maritime Authority \textit{Søfartsstyrelsen} publicly provides a collection of historic AIS data from the region of Denmark \cite{aisdata}.  The system transmits two main types of data: static information (e.g., vessel name, type, and MMSI) and dynamic information (e.g., timestamp, longitude, Latitude, Speed Over Ground (SOG), and Course Over Ground (COG). While AIS data is invaluable, its quality can be heterogeneous. It is frequently affected by irregularities, including shared or spoofed MMSI identifiers, signal dropouts, and physically implausible transmissions (e.g., sudden jumps in location or speed). These inconsistencies necessitate rigorous preprocessing. Furthermore, maritime vessel traffic exhibits seasonal patterns, with private and recreational ships dominating during summer months. To ensure the relevance of our predictive model to global trade and commercial logistics, we focus exclusively on sampled data from the 1st to the 14th of Oct. 2025, selecting a time frame that emphasizes stable and high-volume movements of commercial cargo vessels.

Recent work increasingly applies deep learning to AIS-based vessel trajectory prediction, with Transformer architectures receiving particular attention due to their ability to capture long-range temporal dependencies. 
In addition to architectural innovations, several complementary strategies have been proposed to improve long-term predictions. SEMINT~\cite{chen2025semint} incorporates large language models and historical maritime knowledge. Multi-modal knowledge-enhanced frameworks~\cite{yu2025multi} combine AIS data with environmental and domain information, and probabilistic deep learning approaches~\cite{sorensen2022probabilistic} provide uncertainty-aware forecasts. 
These methods highlight challenges in long-term trajectory prediction that extend beyond standard sequence modeling.
TPTrans~\cite{tptrans} addresses some of these by retaining the standard Transformer encoder–decoder structure and augmenting it with convolutional layers that extract local spatiotemporal features before global attention is applied. While this CNN-Transformer hybrid improves prediction accuracy and captures complex spatial and temporal relationships in AIS data, it inherits several limitations of existing deep learning approaches: the model does not incorporate explicit temporal context features, predictions are generated in absolute rather than relative coordinates, and confidence decreases as the prediction horizon grows.
Similar limitations are observed in TrAISformer~\cite{li2024traisformer}, which employs a spatio-temporal Transformer and demonstrates improvements over recurrent architectures, yet still exhibits increasing error for extended prediction horizons. 

These limitations are compounded by the computational constraints of the vanilla transformer, whose quadratic $L^2$ self-attention cost becomes prohibitive for long input sequences. This motivates the exploration of architectures specifically designed for long-horizon forecasting. The Informer~\cite{informer} architecture addresses these challenges through probabilistic sparse attention and a generative decoder, enabling efficient processing of long sequences while avoiding iterative autoregressive decoding. Across four benchmark datasets, Informer demonstrates stable long-range behavior, with errors increasing smoothly as the horizon widens, and consistently outperforms related Transformer variants such as Reformer \cite{kitaev2020reformer} and LogTrans \cite{nie2022logtrans} as well as LSTMa \cite{bahdanau2014neural} networks. 

Taken together, TPTrans and informers provide complementary perspectives on trajectory prediction: TPTrans highlights the value of combining local convolutional features with attention, while informers show how architectural efficiency and sparse attention can support long-range forecasting. These developments form the basis of our benchmarking and clarify the methodological gaps our work aims to address.

\section{Methodology}
\label{sec:methodology}

\subsection{Data Processing}
For the preprocessing of the AIS trajectory data, the following steps were undertaken:
AIS messages outside the defined geographical study region were removed, and only vessels with valid MMSI identifiers and acceptable mobile classes (A and~B) were retained. For each vessel, the trajectory was ordered by timestamp, duplicate records were removed, and physically implausible tracks were excluded based on minimum duration, maximum allowable SOG, and overall track length.
Trajectories were segmented whenever the temporal gap between consecutive AIS messages exceeded 15 minutes, and segments failing the same quality criteria (minimum duration and motion) were discarded.
All remaining trajectories were resampled to a fixed 6-minute interval and projected from geographic coordinates into Web~Mercator $(x,y)$ coordinates to enable faster distance calculations for filtering.
 Physically inconsistent points were filtered out by enforcing limits on step distance ($\leq 50$~km), time gaps ($\leq 4$~h), and implied speed ($\leq 100$~km/h), while points below a minimum speed threshold of 6~km/h were also removed to exclude near-stationary behaviour.
 Cleaned trajectories were divided into continuous blocks, and a sliding window technique was applied: a 60-minute window was moved with fixed stride, with the first 30 minutes forming the model input and the following 30 minutes serving as prediction targets.

\subsection{Model Architecture}


The proposed architecture TempTPI is closely related to the TPTrans model \cite{tptrans}, but introduces two novel improvements in the form of a temporal input encoding and an Informer encoder. The input is processed in two separate encoder channels and combined before using a positional encoder \cite{attention}. The encoded input goes through a 1d-convolutional layer, followed by an informer encoder layer. The encoded content is mean-pooled and combined with a positional decoder output, before entering separate decoders in the form of fully connected layers for latitude and longitude prediction. The following paragraphs go into more detail on the temporal encoding and the informer layer.

The temporal information transmitted in the AIS data serves as a valuable resource to detect seasonal pattern variations in the vessel behavior. To allow our model to capture these variations, we propose a temporal encoder in the first layer of the model.

\begin{figure}[htbp] 
    \centering
    \includegraphics[width=\linewidth]{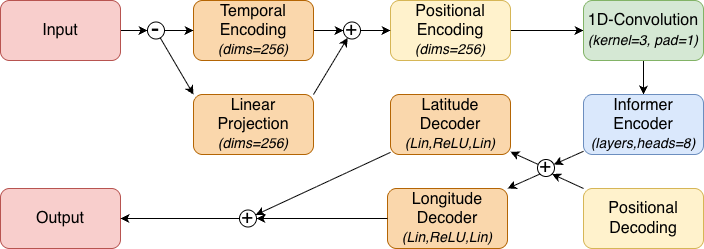}
    \caption{Model Architecture of TempTPI}
    \label{fig:architecture}
\end{figure}

 The dataset was extended to contain sine and cosine cyclic transformations to capture the \textit{hour of day}, \textit{day of week} and \textit{month of year} cycles. The temporal encoder uses a linear projection for each cyclic feature, as well as one bias parameter for each projection. To process a richer representation of the cyclic features \cite{fourier}, we introduce a Fourier-like frequency expansion across a frequency range $K$ such that the temporal encoding $TE$ for each cyclic feature $c \in \{h, d, m\}$ is given as:
\[
TE_c(x)
\begin{cases}
    sin(k*x)\\
    cos(k*x), \text{ for } k = 1,..,K.
\end{cases}
\]
These expanded features are fed into the corresponding projection layer. The encoder then combines all projections and their respective biases. The output of the temporal encoder is combined with the output of the normal data projection layer and fed into the positional encoder.

\begin{figure}[htbp]
    \centering
    \includegraphics[width=1\linewidth]{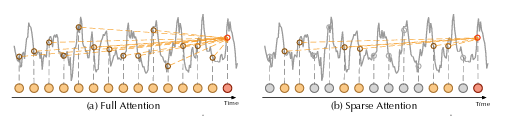}
    \caption{Full vs Sparse Attention \cite{sparseattention}}
    \label{fig:sparseattention}
\end{figure}

TempTPI builds on the \textbf{Informer } architecture, which relies on three main components: ProbSparse self-attention, self-attention distilling, and a generative decoder.
ProbSparse Self-attention that is defined as: 

\begin{equation}
    \mathcal{A}(Q, K, V) = \mathrm{Softmax}\!\left( \frac{\overline{Q}K^\top}{\sqrt{d}} \right) V ,
\end{equation}
where $\overline{Q}$ is a sparse matrix of the same size as $Q$, containing only the top-$u$ queries under the sparsity measurement $M(q, K)$. This allows the model to evaluate only $O(\log L_{Q})$ dot-products for each query, instead of the full $O(L_{Q})$ required in the standard Transformer. The intuition behing this mechanism can be seen on Figure \ref{fig:sparseattention}. 

The Informer also modifies the architecture through self-attention distilling and a 
generative decoder. The encoder compresses the sequence after each ProbSparse 
attention block using convolution and pooling, reducing memory from quadratic 
to roughly linear $O((2-\varepsilon)L\log L)$ and enabling efficient processing of very 
long inputs. The decoder replaces autoregressive decoding with a single-step 
generative formulation, taking a start token and predicting the full horizon at 
once, which removes cumulative error and avoids the $O(L_y)$ iterative cost of 
the standard Transformer \cite{informer}.

%



\subsection{Model Training}

The proposed architecture, TempTPI, was evaluated alongside TPInform (TPTrans, with an Informer encoder instead of a transformer) and TPTrans. The primary goal of the training phase was to establish the effectiveness of the temporal encoder and to quantify the contribution of the ProbSparse self-attention mechanism via an ablation study. Since TempTPI showed consistent improvement to TPInform, TPInform is omitted from the results. 

All models were developed and trained using PyTorch 2.4.1 and executed on GPU with an available 4GB per core. The optimization strategy employed the Adam optimizer \cite{loshchilov2017fixing} with a learning rate $\mu = 1\mathrm{e}{-4}$ and a batch size of 32. Training was executed for a maximum of 200 epochs. Additionally, a standard train-validation-test split ($70\%$, $20\%$, and $10\%$, respectively) was applied to the total dataset. To compare the proposed trajectory prediction model with the aforementioned TPTrans model \cite{tptrans}, we used \textit{Mean Squared Error} (MSE) as the loss function. The loss was calculated with respect to the predicted and actual latitude and longitude as
\begin{equation*}
    MSE = \frac{1}{2N}\sum_{i\in N} [(lon^{(i)}_{true}-lon^{(i)}_{pred})^2+  (lat^{(i)}_{true}-lat ^{(i)}_{pred})^2]
\end{equation*}

\begin{figure}[htbp] 
    \centering
    \includegraphics[width=1.0\linewidth]{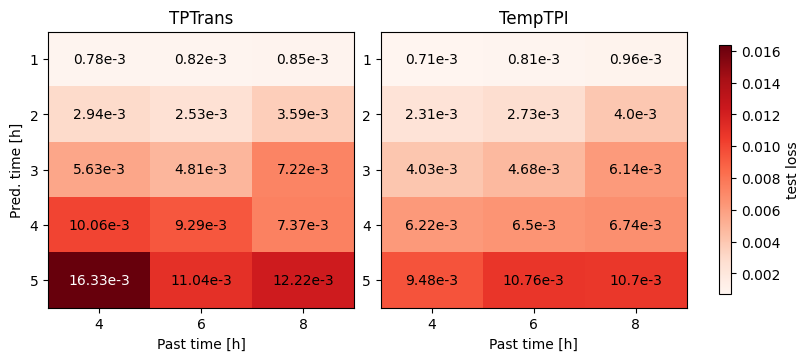}
    \caption{Ablation study results for varying historical context and prediction time}
    \label{fig:block1}
\end{figure}

A systematic ablation study was designed to investigate the model's sensitivity to the data sampling parameters and its impact on the performance quality. The experiments were split into two blocks, targeting the goal of 21 tests for each model. The first block of experiments focused on the impact of the temporal dimensions of the input and prediction sequences. These tests used a full dataset ($k=1000$unique MMSIs of vessels) and a fixed sliding window stride of 30 minutes. The varied parameters were the input window size, providing a historical context across {240, 360, 480} minutes, and the prediction window size, targeting the forecast length varying across {60, 120, 180, 240, 300} minutes. The output of this experiment can be seen in Figure \ref{fig:block1}. While TempTPI shows an improvement, especially for longer predictions, it can also be seen that the performance dropped for longer input lengths. Since the dataset size decreases with increasing input length, we take from this that TempTPI requires more input to achieve good results.

\begin{figure}[htbp] 
    \centering
    \includegraphics[width=1.0\linewidth]{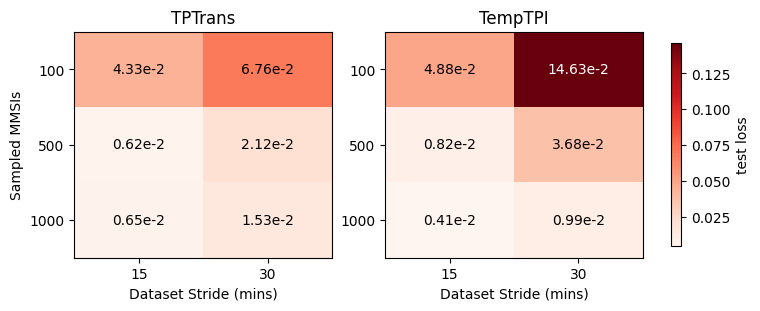}
    \caption{Ablation study results for varying data volume and stride}
    \label{fig:block2}
\end{figure}

To complete the study and evaluate model prediction quality under varying data volume and stride, a second block of 6 experiments was conducted. These tests used the fixed value of the input window size of 480 minutes and the prediction window size of 300 minutes. The varied parameters were the data volume $k$ sampled across {100, 500, 1000} unique MMSIs simulating scenarios with high, moderate, and abundant data. The other tested parameter was sliding window stride that varied across {15, 30} minutes. The impact of this experiment can be seen in Figure \ref{fig:block2}. Since TempTPI performs significantly worse on the smallest sample size, it confirms the previous assumption that the required amount of data is larger than for TPTrans.

\section{Results}
To evaluate the performance of the model, we used the same evaluation metrics as explained in the previous section. The training for these models was conducted on identical training and validation datasets, with an increasing sample number for increasing prediction length, to prevent the underfitting mentioned in the ablation study by keeping a consistent dataset length of approx. 40000 windows. The tests were performed on a separate test dataset, consisting of approximately 8000 trajectory windows with a stride of 15 minutes, which was not part of the model training procedure. For all prediction lengths, a fixed past time window of 5 hours was used. The curve was fitted to a quadratic function $y(x) = ax^2+bx+c$ using \texttt{curve\_fit} \cite{virtanen2020scipy}, to get a generalized estimate of the loss over prediction length. 

\begin{figure}[htbp]
    \centering
    \includegraphics[width=\linewidth]{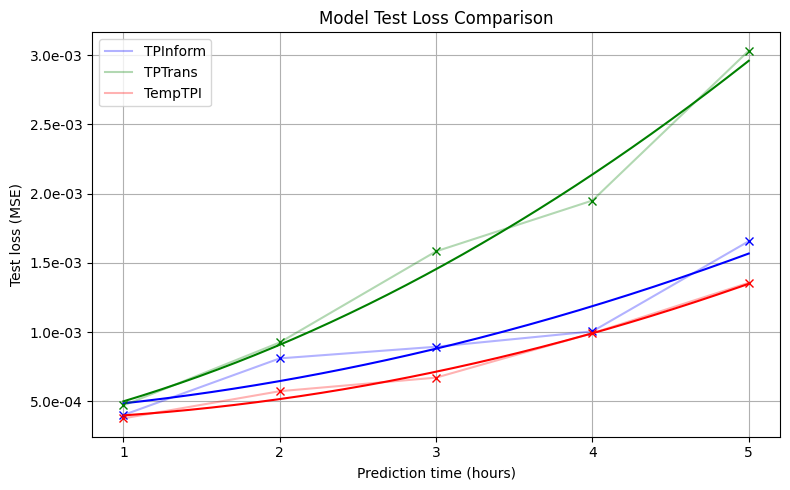}
    \caption{Comparison of loss (MSE) over prediction length}
    \label{fig:loss_comp}
\end{figure}

A comparison of the loss for different prediction lengths can be seen in Fig. \ref{fig:loss_comp}. It can be seen that switching to an informer-based encoder (blue curve) significantly reduces the loss in comparison to the original TPTrans model (green curve). Additionally, adding the temporal encoder to this architecture gives a consistent improvement across all prediction lengths. Most importantly, at the prediction length of 5 hours, TempTPI shows a roughly 55\% improvement over TPTrans.

\begin{figure}[htbp]
    \centering
    \includegraphics[width=\linewidth]{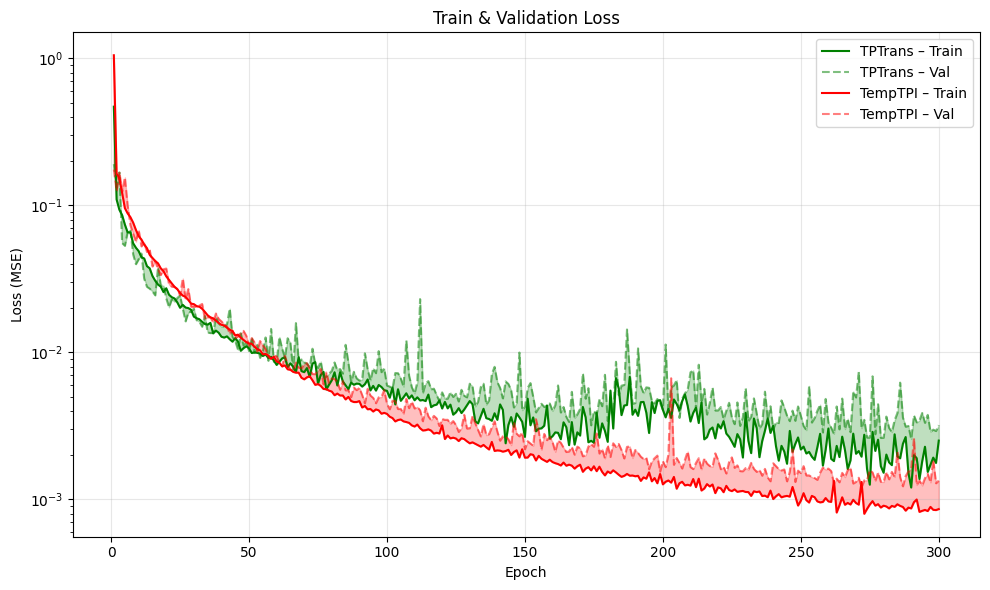}
    \caption{Comparison of train and val. loss (MSE) over epochs}
    \label{fig:train_loss}
\end{figure}

In Fig. \ref{fig:train_loss} we can see the development of the training and validation loss values during the training for the prediction length of 5 hours. The used learning rate was $\mu = 1\mathrm{e}{-4}$ for both models. While TempTPI initially has a higher loss, after around 50 epochs it performs better than TPTrans. It additionally shows a longer and more stable learning curve. 

\begin{figure}[htbp]
    \centering
    \includegraphics[width=\linewidth]{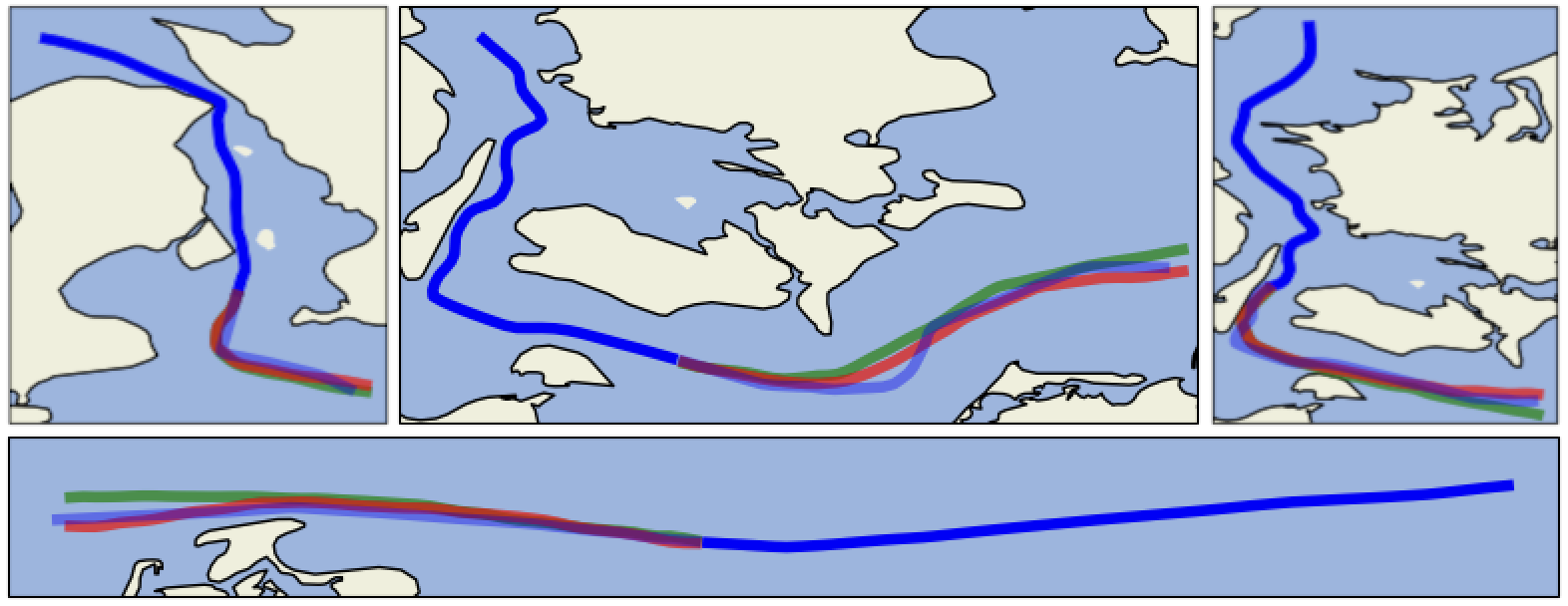}
    \caption{Comparative prediction samples for 3-5 hours. TempTPI (red), actual (light blue), TPTrans (green)}
    \label{fig:samples}
\end{figure}

Looking at some samples of the predicted trajectories for long predictions, shown in Fig. \ref{fig:samples}, one can see that, with the given actual trajectory (blue), the predicted paths of TempTPI (red) are generally closer to the actual paths (light blue) than predictions from TPTrans (green).

We can see a tendency for curves to be underestimated in some and overestimated in other predictions. The predicted trajectories successfully navigate narrow pathways, e.g., in the Øresund and Great Belt straits, relatively well, but sometimes violate water boundaries. 

In general the proposed model is very effective, even compared to a state-of-the-art model such as TPTrans. The significant reduction in test loss for very long predictions suggests greater stability and a higher learning potential, especially regarding cyclic fluctuations in vessel behavior captured with the temporal encoding.

\section{Discussion}

This study shows the potential of using an Informer-based encoder in combination with a temporal encoding, which has been compared to a transformer-based model and yielded significantly better results. The higher accuracy, especially for longer prediction windows, combined with capturing cyclic variances, suggests high potential for usage in the field of maritime logistics. 
The focus of this paper has been on the model architecture, and the layer parameters were kept relatively consistent; experimenting with different parameters could further increase the efficiency.  

However, it has to be mentioned that the used dataset only captured traffic for two weeks in a single month. Since the temporal encoder aims to capture hourly, daily, and monthly variations, the model could benefit from further training on a larger dataset, ensuring that all seasons, such as spring and summer, are well represented and enabling further prediction windows. 
As of right now, only predictions in the area of Denmark have been investigated. To account for varying maritime domains, this work could potentially be extended to other areas of the world, focusing on high-traffic areas such as the Suez Canal or Panama Canal.  

The results have partially shown trajectories being predicted to enter land boundaries. In a novel experiment, we introduced a land punishment during training, but found that it destabilized the training and reduced overall performance, thus it was left out of this paper. Future research could investigate options for penalized training or other land avoidance techniques in more detail.

To summarize, the presented architecture demonstrated a promising foundation for long-term prediction, but its full potential of capturing seasonal variations remains to be investigated in the future. These findings highlight a pathway toward more accurate, reliable, and operationally valuable forecasting models within the global maritime domain.

\small
\bibliographystyle{IEEEtranN}
\bibliography{references}

@inproceedings{informer,
  title        = {Informer: Beyond Efficient Transformer for Long Sequence Time-Series Forecasting},
  author       = {Zhou, H. and Zhang, S. and Peng, J. and Zhang, S. and Li, J. and Xiong, H. and Zhang, W.},
  booktitle    = {Proceedings of the AAAI Conference on Artificial Intelligence},
  volume       = {35},
  number       = {12},
  pages        = {11106--11115},
  year         = {2021},
  url          = {https://doi.org/10.1609/aaai.v35i12.17325}
}

@inproceedings{tptrans,
author = {Wang, W. and Xiong, W. and Ouyang, X. and Chen, L.},
title = {{TPTrans}: Vessel Trajectory Prediction Model Based on Transformer Using {AIS} Data},
booktitle = {ISPRS International Journal of Geo-Information},
volume = {13},
year = {2024},
number = {11},
article-number = {400},
url = {https://doi.org/10.3390/ijgi13110400},
}

@misc{seabornetrade,
  author       = {{United Nations Conference on Trade and Development (UNCTAD)}},
  title        = {Shipping data: seaborne trade statistics},
  year         = {2025},
  howpublished = {\url{https://unctad.org/news/shipping-data-unctad-releases-new-seaborne-trade-statistics}},
  note         = {Accessed: 2025-11-23}
}

@article{benefits,
author = {Li, X. and Liu, C and Li, J. and Zhao, L. and Du, Z.},
title = {Advancing ship trajectory prediction: Integrating deep learning with enhanced reference trajectory correction techniques},
journal = {Ocean Engineering},
volume = {311},
pages = {118880},
year = {2024},
url = {https://doi.org/10.1016/j.oceaneng.2024.118880},
}

@misc{aisdata,
author = {Danish Maritime Authority - Søfartsstyrelsen},
title = {{AIS}-data},
note = {Accessed: 2025-11-23},
howpublished = {\url{https://www.soefartsstyrelsen.dk/sikkerhed-til-soes/sejladsinformation/ais-data}}
}

@article{reviewofais,
  author    = {Felski, A. and Jask{\'o}lski, K. and Bany{\'s}, P.},
  title     = {Comprehensive Assessment of Automatic Identification System {(AIS)} Data in Regard to Vessel Movement Prediction},
  journal   = {The Journal of Navigation},
  volume    = {67},
  number    = {5},
  pages     = {791--809},
  year      = {2015},
  doi       = {10.1017/S0373463314000253},
  url       = {https://doi.org/10.1017/S0373463314000253}
}

@inproceedings{fourier,
 author = {Matthew Tancik et al.},
 booktitle = {Advances in Neural Information Processing Systems},
 editor = {H. Larochelle and M. Ranzato and R. Hadsell and M.F. Balcan and H. Lin},
 pages = {7537--7547},
 publisher = {Curran Associates, Inc.},
 title = {Fourier Features Let Networks Learn High Frequency Functions in Low Dimensional Domains},
 url = {https://proceedings.neurips.cc/paper_files/paper/2020/file/55053683268957697aa39fba6f231c68-Paper.pdf},
 volume = {33},
 year = {2020}
}

@inproceedings{attention,
 author = {Vaswani, Ashish and Shazeer, Noam and Parmar, Niki and Uszkoreit, Jakob and Jones, Llion and Gomez, Aidan N and Kaiser, \L ukasz and Polosukhin, Illia},
 booktitle = {Advances in Neural Information Processing Systems},
 editor = {I. Guyon and U. Von Luxburg and S. Bengio and H. Wallach and R. Fergus and S. Vishwanathan and R. Garnett},
 pages = {},
 publisher = {Curran Associates, Inc.},
 title = {Attention is All you Need},
 url = {https://proceedings.neurips.cc/paper_files/paper/2017/file/3f5ee243547dee91fbd053c1c4a845aa-Paper.pdf},
 volume = {30},
 year = {2017}
}

@article{sparseattention,
  author       = {Haixu Wu and
                  Jiehui Xu and
                  Jianmin Wang and
                  Mingsheng Long},
  title        = {Autoformer: Decomposition Transformers with Auto-Correlation for Long-Term
                  Series Forecasting},
  journal      = {CoRR},
  volume       = {abs/2106.13008},
  year         = {2021},
  url          = {https://arxiv.org/abs/2106.13008},
  eprinttype    = {arXiv},
  eprint       = {2106.13008},
  bibsource    = {dblp computer science bibliography, https://dblp.org}
}

@article{chen2025semint,
  title={{SEMINT}: an {LLM}-empowered long-term vessel trajectory prediction framework},
  author={Chen, Nanyu and Yang, Anran and Wu, Hui and Chen, Luo and Xiong, Wei and Jing, Ning},
  journal={International Journal of Geographical Information Science},
  pages={1--35},
  year={2025},
  publisher={Taylor \& Francis}
}

@inproceedings{yu2025multi,
  title={A Multi-Modal Knowledge-Enhanced Framework for Vessel Trajectory Prediction},
  author={Yu, Haomin and Li, Tianyi and Torp, Kristian and Jensen, Christian S},
  booktitle={Proceedings of the 19th International Symposium on Spatial and Temporal Data},
  pages={44--54},
  year={2025}
}

@inproceedings{li2024traisformer,
  title={{TrAISformer}: Spatio-Temporal Ship Trajectory Prediction Based on Transformer},
  author={Li, Yunbo and Wang, Jiayu and Li, Tao and Fu, Zheng},
  booktitle={2024 5th International Seminar on Artificial Intelligence, Networking and Information Technology (AINIT)},
  pages={1099--1104},
  year={2024},
  organization={IEEE}
}

@article{sorensen2022probabilistic,
  title={Probabilistic maritime trajectory prediction in complex scenarios using deep learning},
  author={S{\o}rensen, Kristian Aalling and Heiselberg, Peder and Heiselberg, Henning},
  journal={Sensors},
  volume={22},
  number={5},
  pages={2058},
  year={2022},
  publisher={MDPI}
}

@article{kitaev2020reformer,
  title={Reformer: The efficient transformer},
  author={Kitaev, Nikita and Kaiser, {\L}ukasz and Levskaya, Anselm},
  journal={arXiv preprint arXiv:2001.04451},
  year={2020}
}

@inproceedings{nie2022logtrans,
  title={Logtrans: Providing efficient local-global fusion with transformer and cnn parallel network for biomedical image segmentation},
  author={Nie, Xingqing and Zhou, Xiaogen and Li, Zhiqiang and Wang, Luoyan and Lin, Xingtao and Tong, Tong},
  booktitle={2022 IEEE 24th Int Conf on High Performance Computing \& Communications; 8th Int Conf on Data Science \& Systems; 20th Int Conf on Smart City; 8th Int Conf on Dependability in Sensor, Cloud \& Big Data Systems \& Application (HPCC/DSS/SmartCity/DependSys)},
  pages={769--776},
  year={2022},
  organization={IEEE}
}

@article{bahdanau2014neural,
  title={Neural machine translation by jointly learning to align and translate},
  author={Bahdanau, Dzmitry and Cho, Kyunghyun and Bengio, Yoshua},
  journal={arXiv preprint arXiv:1409.0473},
  year={2014}
}

@article{loshchilov2017fixing,
  title={Fixing weight decay regularization in adam},
  author={Loshchilov, Ilya and Hutter, Frank and others},
  journal={arXiv preprint arXiv:1711.05101},
  volume={5},
  number={5},
  pages={5},
  year={2017}
}

@article{virtanen2020scipy,
  title={{SciPy} 1.0: fundamental algorithms for scientific computing in Python},
  author={Virtanen, Pauli and Gommers, Ralf and Oliphant, Travis E and Haberland, Matt and Reddy, Tyler and Cournapeau, David and Burovski, Evgeni and Peterson, Pearu and Weckesser, Warren and Bright, Jonathan and others},
  journal={Nature methods},
  volume={17},
  number={3},
  pages={261--272},
  year={2020},
  publisher={Nature Publishing Group US New York}
}

@article{shiptracking,
title = {Maritime surveillance: Tracking ships inside a dynamic background using a fast level-set},
journal = {Expert Systems with Applications},
volume = {38},
number = {6},
pages = {6669-6680},
year = {2011},
issn = {0957-4174},
doi = {https://doi.org/10.1016/j.eswa.2010.11.068},
url = {https://www.sciencedirect.com/science/article/pii/S0957417410013060},
author = {Zygmunt L. Szpak and Jules R. Tapamo},
}

@article{maritimescenario,
  author       = {Dilip K. Prasad and
                  Chandrashekar Krishna Prasath and
                  Deepu Rajan and
                  Lily Rachmawati and
                  Eshan Rajabally and
                  Chai Quek},
  title        = {Challenges in video based object detection in maritime scenario using
                  computer vision},
  journal      = {CoRR},
  volume       = {abs/1608.01079},
  year         = {2016},
  url          = {http://arxiv.org/abs/1608.01079},
  eprinttype   = {arXiv},
  eprint       = {1608.01079},
  bibsource    = {dblp computer science bibliography, https://dblp.org}
}

\end{document}